\documentclass[10pt,twocolumn,letterpaper]{article}

\usepackage{wacv}              

\usepackage{graphicx}
\usepackage{amsmath}
\usepackage{amssymb}
\usepackage{booktabs}

\usepackage[accsupp]{axessibility} 

\usepackage{multirow}

\usepackage{adjustbox}

\usepackage[pagebackref,breaklinks,colorlinks]{hyperref}

\usepackage[capitalize]{cleveref}
\crefname{section}{Sec.}{Secs.}
\Crefname{section}{Section}{Sections}
\Crefname{table}{Table}{Tables}
\crefname{table}{Tab.}{Tabs.}

\def\wacvPaperID{216} 
\def\confName{WACV}
\def\confYear{2024}

\begin{document}

\title{CVTHead: One-shot Controllable Head Avatar with Vertex-feature Transformer}


\author{Haoyu Ma, \hspace{1pt} Tong Zhang, \hspace{1pt} Shanlin Sun, \hspace{1pt} Xiangyi Yan, \hspace{1pt} Kun Han,  \hspace{1pt} Xiaohui Xie \\
University of California, Irvine \\
\{haoyum3, tongz27, shanlins, xiangyy4, khan7, xhx\}@uci.edu}

\maketitle

\begin{abstract}
Reconstructing personalized animatable head avatars has significant implications in the fields of AR/VR. Existing methods for achieving explicit face control of 3D Morphable Models (3DMM) typically rely on multi-view images or videos of a single subject, making the reconstruction process complex. Additionally, the traditional rendering pipeline is time-consuming, limiting real-time animation possibilities.
In this paper, we introduce CVTHead, a novel approach that generates controllable neural head avatars from a single reference image using point-based neural rendering. CVTHead considers the sparse vertices of mesh as the point set and employs the proposed Vertex-feature Transformer to learn local feature descriptors for each vertex. This enables the modeling of long-range dependencies among all the vertices. Experimental results on the VoxCeleb dataset demonstrate that CVTHead achieves comparable performance to state-of-the-art graphics-based methods. Moreover, it enables efficient rendering of novel human heads with various expressions, head poses, and camera views. These attributes can be explicitly controlled using the coefficients of 3DMMs, facilitating versatile and realistic animation in real-time scenarios. Codes and pre-trained model can be found at \href{https://github.com/HowieMa/CVTHead}{https://github.com/HowieMa/CVTHead}. 

\end{abstract}

\section{Introduction}
Personalized head avatars play a crucial role in a wide range of applications, including AR/VR, teleconferencing, and the movie industry. Over the past few decades, there has been an extensive exploration of personalized head avatars in the fields of computer graphics and computer vision.
Traditional solution \cite{alexander2010digital} reconstructs a personalized mesh and texture for the source actor explicitly with 3D head scans \cite{yang2020facescape,wuu2022multiface}. 
To perform full face control, 3D Morphable Models (3DMM) \cite{blanz1999morphable,li2017learning} are used as a strong prior of face geometry. 
3DMMs is a parametric model and uses PCA-based linear blendshapes to explicitly control face shape, expressions, texture, and head pose independently. However, 3DMM does not model the facial detail and hair region of the human face \cite{grassal2022neural}. 
Recently, with the development of Neural Radiance Fields (NeRF) \cite{mildenhall2020nerf}, reconstructing avatars with implicit models becomes popular as it can reconstruct detailed regions \cite{gafni2021dynamic,park2021nerfies,athar2022rignerf}. 
However, all these methods are subject-specific and they usually require video inputs or multi-view images of the same subject, which limits their usage in practice.

Hence, acquiring human avatars from a single image (i.e., one-shot face reenactment) becomes more and more popular \cite{wiles2018x2face,thies2020neural,zhang2023sadtalker,siarohin2019first,zakharov2020fast,wang2021one,thies2016face2face,yao2020mesh,doukas2021headgan,khakhulin2022realistic}. Given a facial image of an actor, the synthetic images can be driven by videos from other actors. A key step behind these methods is to decouple the facial appearance and motion information from the source and driven images. 
As a result, mesh-guided face animation has gained significant attention, primarily due to the inherent disentanglement of identity and expression offered by 3DMM.
Generally, one-shot mesh-guided face animation can be roughly divided into warp-based and graphics-based. 
Warp-based methods \cite{yao2020mesh,doukas2021headgan,yang2022face2face,zeng2022fnevr} employ the motion field to transfer the driving pose and expression into the source face. 
These methods effectively preserve fine facial details and produce high-fidelity results but only work well for a limited range of head poses. 
Graphic-based methods \cite{feng2021learning,khakhulin2022realistic} learn texture maps \cite{blinn1976texture} from single-image and apply computer graphics pipelines to render the animated face image. Thus, it can maintain performance under large head rotations and guarantee the 3D consistency of rendered images. 
However, the rendering pipeline is usually computationally heavy \cite {kato2020differentiable}, which makes efficient rendering unachievable.

Alternatively, point-based graphics \cite{gross2011point} get rid of the surface mesh and directly use point clouds to model the 3D geometry. Later on, the point-based neural rendering techniques \cite{aliev2020neural,rakhimov2022npbg++,xu2022point} augment each RGB point with a learnable neural descriptor that is interpreted by the neural renderer. The recent SMPLpix \cite{prokudin2021smplpix} further extends these techniques from static scenes to dynamic scenes and enables the efficient rendering of human body avatars under novel subject identities and human poses. Although efficient, these methods still require multi-view images with calibrated cameras to reconstruct the accurate point cloud first. 
Consequently, applying these methods to one-shot face reenactment is infeasible.

In this paper, we utilize point-based neural rendering to achieve an efficient and realistic generation of head avatars from a single image. 
We direct utilize the sparse vertices from the FLAME head model \cite{li2017learning} as our point set, instead of reconstructing a dense point cloud of the subject tediously.
Specifically, given the vertices from pre-trained 3D face reconstruction networks \cite{feng2021learning}, we learn a local feature descriptor aligned with each vertex. 
When learning the local descriptor of a 3D point from a 2D reference image, pixel-aligned features \cite{saito2019pifu,huang2020arch,he2021arch++,corona2023structured} are a popular choice. However, these aligned features often become incorrect when the projected 2D location is occluded in the source image. 
To address this challenge, we propose the \textit{Vertex-feature Transformer}. This approach treats each vertex as a query token \cite{dosovitskiy2021an} and utilizes transformers \cite{vaswani2017attention} to directly learn the canonical vertex features from the reference image. By incorporating a global attention mechanism, our model can capture long-range dependencies within the features of all vertices. Thus, the feature descriptor of invisible 3D points can still be reconstructed correctly. 
Next, we project the feature descriptor and depth of each vertex into image space and employ a UNet-like neural rendering to generate the RGB image. 
Since the feature descriptor is aligned with the vertices of the FLAME head model, the rendered face can be explicitly controlled by the shape, expression, and head pose coefficients. We name this end-to-end framework as CVTHead, in short for Controllable head avatar with Vertex-feature Transformer.

Our major contributions are summarized as follows:
\begin{itemize}
    \item We propose CVTHead, a one-shot controllable head avatar framework using point-based neural rendering that can efficiently render novel human heads under novel expressions and camera views. To the best of our knowledge, this is the first work that performs point-based neural rendering from a monocular face image. 

    \item We propose Vertex-feature Transformer to learn the vertex descriptor in canonical space from a single image with transformers, and demonstrate its superiority beyond projection methods. 

    \item  By conducting experiments on VoxCeleb1 and VoxCeleb2, we establish that our method achieves performance that is on par with the state-of-the-art approaches, while additionally improving efficiency.
    
\end{itemize}

\section{Related Work}

\paragraph{Mesh-guided Face Reenactment}
Extensive research has been conducted on employing 3DMM for the explicit animation of human face images \cite{tewari2020stylerig,ghosh2020gif,yao2020mesh,doukas2021headgan,wang2021safa,yang2022face2face,khakhulin2022realistic, Drobyshev22MP,zeng2022fnevr}. 
Mesh-guided face reenactment can be divided into warp-based and graphic-based. 
Warping-based methods \cite{yao2020mesh,doukas2021headgan,wang2021safa,yang2022face2face} warp the source image with explicit motion fields. 
For example, given both source and driving meshes, Yao et al. \cite{yao2020mesh} extract the motion features with Graph Convolutional Networks.  HeadGAN \cite{doukas2021headgan} learns the dense flow field with PNCC \cite{zhu2016face} and SPADE. Face2Face$^\rho$ \cite{yang2022face2face} calculates the motion with a set of pre-specified 3D keypoints. 
However, when faced with significant head rotations, the quality of these approaches drops significantly. 
Meanwhile, other methods \cite{feng2021learning,khakhulin2022realistic,grassal2022neural} obtain the animated face images from the head mesh with the classic graphics rendering pipeline. In detail, DECA \cite{feng2021learning} simultaneously learns both the head mesh and the linear albedo subspace of the Basel Face Model \cite{paysan20093d}. 
To create realistic face photos, ROME \cite{khakhulin2022realistic} estimates a neural texture and offset for each vertex from the source image and renders the rigged mesh with deferred neural rendering technique \cite{thies2019deferred}. 
Nevertheless, these methods still require the time-consuming classic differentiable rendering \cite{ravi2020accelerating}.

\begin{figure*}[!t]
    \centering
    \includegraphics[width=0.95 \linewidth]{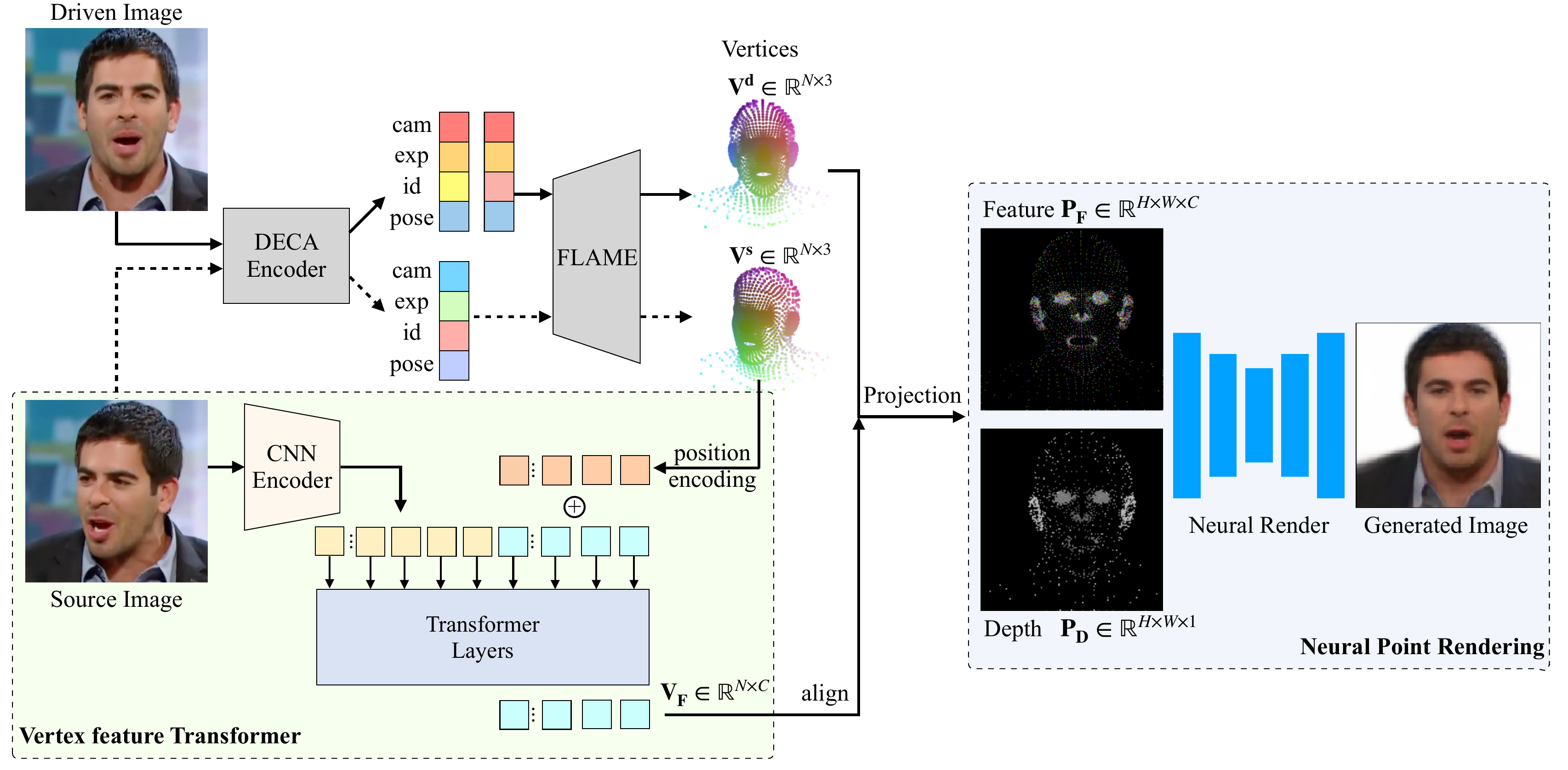}
    \caption{\small{Overview of the CVTHead framework. We employ a pretrained face reconstruction network \cite{feng2021learning} to obtain the face mesh (Section \ref{sec:mesh_recon}) and utilize the proposed verte feature transformers to obtain the feature descriptor of each vertex from the source image (Section \ref{sec:vtx_transformer}). We then consider the sparse vertex as point set and use point-based neural rendering to synthesize the image (Section \ref{sec:neural_rendering}).   }  }
    \label{fig:overview}
\end{figure*}

\paragraph{Neural Head Avatars}
Recently, several works extend NeRF \cite{aliev2020neural} to model dynamic objects such as virtual avatars with implicit neural representations \cite{gafni2021dynamic,park2021nerfies,kania2022conerf,athar2022rignerf,hong2022headnerf,zhuang2022mofanerf}. For example, subject-dependent methods such as NerRACE \cite{gafni2021dynamic} and RigNeRF \cite{athar2022rignerf} use 3DMM-guided deformation neural fields to enable control over head pose, facial expression, and viewpoints. Typically, these approaches employ an optimization-based head tracker \cite{thies2016face2face} as a preprocessing step to extract accurate 3DMM coefficients from a monocular video of the subject. Meanwhile, subject-agnostic methods such as HeadNeRF \cite{hong2022headnerf} and MofaNeRF \cite{zhuang2022mofanerf} learn the radiance fields from large-scale multi-view images \cite{yang2020facescape}. When generating an avatar for a novel subject, these methods require time-consuming inverse rendering optimization to obtain the latent codes. 
Most recently, HiDe-NeRF \cite{li2023one} and OTAvatar \cite{ma2023otavatar} employ tri-plane representations \cite{chan2022efficient} to efficiently extract multi-scale features for each query 3D point and also use volume rendering to reconstruct images. Although promising, a limitation of volumetric rendering is the necessity to sample hundreds of 3D points per ray and then feed them through the network to render a single pixel or feature patch. In contrast, our method utilizes a set of points to represent the head avatar, thereby necessitating only a single forward pass for rendering.

\paragraph{Neural Point-based Rendering}
Neural point-based rendering \cite{aliev2020neural,kopanas2021point,rakhimov2022npbg++,xu2022point,kerbl3Dgaussians} has gained significant attention in recent years for its ability to generate high-quality images by directly rendering point clouds from static scenes. Aliev et al. \cite{aliev2020neural} introduced Neural Point-Based Graphics (NPBG), which employs learnable neural descriptors to enhance each point for better rendering. Later on, NPBG++ \cite{rakhimov2022npbg++} further predicts the descriptors with a single pass to accelerate rendering. 
Meanwhile, SMPLpix \cite{prokudin2021smplpix} extends point-based neural rendering to generate human avatars under the control of SMPL \cite{loper2015smpl}. Although SMPLpix is a subject-agnostic method, it requires registering the SMPL model to ground truth 3D scans to obtain the RGB color of each vertex. 
The most recent PointAvatar \cite{zheng2023pointavatar} models the human head as an explicit canonical point cloud and continuous deformation to create realistic and relightable head avatars. However, it is still subject-dependent and requires a video caption of the subject to train the model. Our method also employs neural point-based rendering but simplifies the setting as we only require a single image of the novel subject.

\paragraph{Transformers in Mesh}
Over the past few years, transformers \cite{vaswani2017attention} have made significant progress in many computer vision tasks \cite{touvron2020training,dosovitskiy2021an,carion2020end, yan2022after,ma2022ppt,yang2023capturing}. 
There are a few works that also apply transformers to mesh data  \cite{lin2020end,lin2021mesh,cho2022FastMETRO,yoshiyasu2023deformable,zheng2023potter,dou2023tore}. In detail, METRO \cite{lin2020end} apply transformers to predict the mesh coordinates and 3D joints simultaneously of the human body \cite{loper2015smpl}. 
Mesh Graphormer \cite{lin2021mesh} further utilizes the topology of the mesh with graph convolution to improve the mesh reconstruction. Both works consider each vertex as a query token and use transformers to learn the non-local relationships among vertices. 
In our work, we also use transformers to learn the correspondence among vertex features.

\section{Methodology}
\vspace{-0.2em}
\subsection{Overview}
\vspace{-0.2em}

Fig \ref{fig:overview} illustrates the overall framework of our CVTHead. Given both source image $\mathbf{I^s}$ and driven image $\mathbf{I^d}$, we utilize a pre-trained face reconstruction model \cite{feng2021learning} to obtain the source and driven vertex coordinates $\mathbf{V^s}$ and $\mathbf{V^d} \in \mathbb{R}^{N \times 3} $ of the FLAME  model \cite{li2017learning}  (Sec. \ref{sec:mesh_recon}), where $N$ is the number of vertices in FLAME model.
Simultaneously, we employ the proposed vertex feature transformer to learn the feature descriptor for all vertices in canonical space $\mathbf{V_F} \in \mathbb{R}^{N \times C}$ from the source image (Sec. \ref{sec:vtx_transformer}), where $C$ is the number of channels of feature descriptor. 
Then we project driven vertices and their corresponding feature descriptors onto the vertex feature image $\mathbf{P_F^d} \in \mathbb{R}^{H \times W \times C}$ and the depth image $\mathbf{P_D^d} \in \mathbb{R}^ {H \times W \times 1}$, where $H$ and $W$ is the height and width of the original image. Next, we conduct neural rendering with a  U-Net $\mathcal{G(\cdot)}$ to generate the synthetic image $\mathbf{\hat{I}} = \mathcal{G}(\mathbf{P_F}, \mathbf{P_D}) \in \mathbb{R}^{H \times W \times 3}$ (Sec. \ref{sec:neural_rendering}). 
Our framework enables end-to-end training, allowing the entire process to be optimized jointly. During inference, our system enables the rendered image to be animated with novel shapes, expressions, head poses, and viewpoints by manipulating the FLAME parameters. This flexibility allows for the generation of diverse and customizable head avatars.

\subsection{Head Mesh Reconstruction}
\label{sec:mesh_recon}

FLAME \cite{li2017learning} is a parametric 3D head model with $N=5023$ vertices. It encompasses a mean template $V_{b} \in \mathbb{R}^{N \times 3}$, along with shape blendshapes $ \mathcal{S} \in \mathbb{R}^{N \times 3 \times L}$, and expression blendshapes $ \mathcal{E} \in \mathbb{R}^{N \times 3 \times K}$.
These blendshapes are derived from a vast collection of 4D scans of human heads, allowing FLAME to capture a wide range of facial variations. 
Given parameters of facial identity $\beta \in \mathbb{R}^L$, expression $\phi \in \mathbb{R}^K$ and pose $\theta \in \mathbb{R}^{3k+3}$ (with $k=4$ joints for neck, jaw, and eyeballs), FLAME first apply $\beta$ and $\phi$ to corresponding blendshapes, resulting in modified vertex positions. 
Next, the linear blend skinning (LBS) technique $W(\cdot, \cdot)$ is employed to rotate the vertices based on $\theta$. The final reconstruction of FLAME in world coordinates is calculated by: 
\begin{equation}
    M(\beta, \phi, \theta) = W(V_b +  \mathcal{S} \beta +  \mathcal{E} \phi, \theta) \in \mathbb{R}^{3n}
\end{equation}

We employ the pre-trained DECA \cite{feng2021learning} $f_D(\cdot)$ to obtain $\beta$,$\phi$,$\theta$ and camera parameters $c$ from both source images and driven images with a single forward, i.e, $\beta^s, \phi^s, \theta^s, c^s = f_D( \mathbf{I^s} )$ and $\beta^d, \phi^d, \theta^d, c^d = f_D(\mathbf{I^d})$. 
We also obtain the deformation of hair and shoulder regions from the source image with the pre-trained linear deformation model $f_H(\cdot)$ \cite{khakhulin2022realistic} to refine the vertices locations. 
Then we obtain the driven vertex coordinates by
\begin{equation}
    \mathbf{V^d} =  M(\beta^s, \phi^d, \theta^d) + f_H(\mathbf{I^s}) \in \mathbb{R}^{N \times 3}
\end{equation}

\subsection{Vertex-feature Transformer}
\label{sec:vtx_transformer}

\paragraph{Motivations}
In previous approaches that utilize pixel-aligned features \cite{saito2019pifu,corona2023structured}, the feature descriptor of a given 3D point is determined by the feature located at its corresponding 2D projection. In detail, given the 3D point $\mathbf{k}^s \in \mathbf{V^s}$, we project it into the 2D image space by $(u^s, v^s, d^s) = \Pi(\mathbf{k}^s, c_s)$, where $\Pi(\cdot)$ represents the orthographic projection function and $c_s$ is the camera parameters of the reference image obtained from pre-trained DECA. The descriptor of $\mathbf{k}^s$ is defined as $I'[u^s, v^s]$, where $I'$ is the 2D feature map of the source image. However, these methods have several limitations. 
First, it requires accurate mesh reconstruction to locate the correct 2D pixels. Moreover, when the point is invisible, the feature at the 2D projection cannot represent the real features of that point. For instance, if the ear is occluded by the face, the projection may result in capturing features from the eye or nose instead. As a result, relying solely on the feature at the 2D projection can lead to incomplete or misleading feature descriptors.

\paragraph{Vertex Feature as Tokens}
To tackle the aforementioned problem, we propose a solution wherein we treat each vertex as an individual query token and leverage the attention mechanism of transformers to acquire its corresponding features from the image feature tokens. This approach avoids the need for a fixed 2D projection and allows for more flexible learning. 
Specifically, we employ $N$ learnable embedding vectors $\mathbf{X_v} \in \mathbb{R}^{N \times C'}$ to represent the feature descriptors associated with each vertex in canonical space and name it as \textit{Vertex Tokens}, where $C'$ is the number of channels.  
To further encode the location information of each vertex, we incorporate the sine positional encoding \cite{vaswani2017attention} to its corresponding image space coordinates $(u^s, v^s)$ and depth $d^s$, denoting as $\mathbf{E^s_{uv}}$ and $\mathbf{E^s_{dep}}$, respectively. 
Finally, the vertex query token is defined as $\mathbf{\Tilde{X}_v} = \mathbf{X_v}+ \mathbf{E^s_{uv}} + \mathbf{E^s_{dep}}$. 
On the other hand, we train a CNN encoder $\mathcal{E}(\cdot)$ to extract feature maps from the source image $\mathbf{I^s}$ and flatten the 2D features into a sequence of tokens $\mathbf{F^s} = \mathcal{E}(\mathbf{I^s}) \in \mathbb{R}^{hw \times C' }$. We also apply the 2D sine positional encodings \cite{dosovitskiy2021an} to encode spatial information, denoted as $\mathbf{E}$. Finally, the image token is defined as $\mathbf{X^s_F} = \mathbf{F^s} + \mathbf{E}$.

\paragraph{Transformers}

The input to the transformer is the concatenation of both image tokens $\mathbf{X^s_F}$ and vertex tokens $\mathbf{\Tilde{X}_v}$, i.e, $\mathbf{X} = [\mathbf{\Tilde{X}_v}, \mathbf{X^s_F}] \in \mathbb{R}^{(N+hw) \times C'}$. 
The standard transformer encoder layer \cite{vaswani2017attention} consists of alternating layers of the multi-headed self-attention (MHSA) and multi-layer perceptron (MLP). 
First, three linear projections are applied to transfer $\mathbf{X}$ into three matrices of equal size, namely the query $\mathbf{Q}$, the key $\mathbf{K}$, and the value $\mathbf{V}$.  
 The self-attention  is calculated by:
\begin{equation}
  \text{SA}(\mathbf{X}) =  \text{Softmax}( \frac{ \mathbf{Q} \mathbf{K}^T }{\sqrt{D}}  )\mathbf{V},
  \label{eq:attention}
\end{equation}
For MHSA, $H$ self-attention modules are applied to $\mathbf{X}$ separately, and each of them produces an output sequence. We utilize the state of the vertex tokens at the output of the transformer encoder and employ a linear transformation to modify its dimensionality, thereby acquiring the vertex descriptor $\mathbf{V_F} \in \mathbb{R}^{N \times C}$.

The vertex-feature transformer has several benefits. Firstly, it eliminates the need for a fixed 2D projection to determine the corresponding feature for each vertex. Instead, it leverages attention mechanisms to identify the relevant feature, introducing a higher degree of flexibility. The transformer incorporates positional encoding to encode location information, further enhancing its adaptability and versatility.
Additionally, the global attention mechanism of transformers facilitates long-range correspondence among all vertex features. Even when the projection of a vertex is occluded, the vertex feature can still be obtained from neighboring regions or symmetrical vertices.

\subsection{Neural Vertex Rendering}
\label{sec:neural_rendering}
Given the learned vertex feature $\mathbf{V_F}$, we further use neural point-based rendering to generate synthetic images. 
During the training, we use the driven vertex $\mathbf{V}^d$ to reconstruct the driven image. In detail, we first project the driven vertices $ \mathbf{k}^d \in \mathbf{V^d}$ into image space with the driven camera parameter $c^d$, i.e., $(u^d,v^d, d^d) = \Pi (\mathbf{k}^d, c^d)$. 
Subsequently, we create the vertex projection features $ \mathbf{P_F^d}  \in \mathbb{R}^{H \times W \times C}$. For each vertex $\mathbf{k}^d$, along with its corresponding descriptor $\mathbf{v_F} \in \mathbb{R}^{C}$, we assign the descriptor to location $(u^d,v^d)$ in the vertex projection features \cite{prokudin2021smplpix}:
\begin{equation}
    \mathbf{P_F^d}[ \lfloor u^d \rfloor, \lfloor v^d \rfloor ] = \mathbf{v_F}
\end{equation}
We keep the features of the nearest vertex when two vertices are projected into the same pixel on $\mathbf{P_F^d}$. For all pixels without projection (i.e., the background pixel), we assign a constant value. Similarly, we also project the depth $d^d$ value into a depth image $P_D$ which satisfies $\mathbf{P_D^d}[ \lfloor u^d \rfloor, \lfloor v^d \rfloor ] = d^d$. 
Finally, we concatenate $\mathbf{P_F^d}$ and $\mathbf{P_D^d}$ and employ a U-Net $\mathcal{G}(\cdot)$ to generate the synthetic image $\hat{\mathbf{I^d}}$ as well as the binary foreground mask $\hat{\mathbf{M^d}}$,  i.e., 
\begin{equation}
    (\hat{\mathbf{I^d}}, \hat{\mathbf{M^d}}) = \mathcal{G}([\mathbf{P_F^d}, \mathbf{P_D^d}]).
\end{equation}

\subsection{Training}
During the training time, we randomly sample $\mathbf{I^s}$ and $\mathbf{I^d}$ from the same video. 
We fixed the pre-trained DECA and only update the parameters of vertex-feature transformers and neural render. 
Following \cite{khakhulin2022realistic}, we use the L1 loss $L_{L1}$, VGG perceptual loss $L_{vgg}$ \cite{johnson2016perceptual}, face recognition loss $L_{id}$ \cite{cao2018vggface2}, and adversarial loss \cite{NIPS2014_5ca3e9b1, wang2018high} $L_{a}$ to measure the difference between the reconstructed driven image $\hat{\mathbf{I^d}}$ and the ground truth $\mathbf{I^d}$. We use the Dice loss to match the predicted segmentation masks. The total loss is calculated by: 
\begin{equation}
    L = \lambda_{L1}  L_{L1} + \lambda_{vgg}  L_{vgg} + \lambda_{id}  L_{id} + \lambda_{seg} L_{seg} + \lambda_{a}  L_{a}
\end{equation}
, where $\lambda_{L1}$, $\lambda_{vgg}$, $\lambda_{id}$, $\lambda_{seg}$ and $\lambda_{a}$ is the corresponding weights of each loss term.

\section{Experiments}

\subsection{Experimental Set up}

\paragraph{Dataset} 
For a fair comparison with previous works, we conduct experiments on VoxCeleb1 \cite{nagrani2017voxceleb} and VoxCeleb2 \cite{chung2018voxceleb2}. 
VoxCeleb1 contains around 20k video sequences of over 1000 actors and VoxCeleb2 contains around 150k videos of over 6000 actors. 
Note that, ROME \cite{khakhulin2022realistic} carefully selects a subset of around 15k high-quality video sequences from VoxCeleb2 for training and evaluation, which is not publicly available. We directly use all VoxCeleb2 videos instead. 
Following \cite{siarohin2019first}, each frame is cropped into $256 \times 256$ and normalized to $[-1, 1]$. 
We follow the identity-based split thus all subjects in the validation set are unseen by the model. 
Besides, we apply an off-the-shelf face parsing network \cite{yu2018bisenet} to obtain the foreground mask of each frame, which is considered as the pseudo ground truth.

\paragraph{Implementation Details}
We use the same CNN encoder $\mathcal{E}(\cdot)$ as in ROME \cite{khakhulin2022realistic}, which downsamples $16\times$ of the original image.  
Naturally, our vertex-feature transformer is able to process arbitrary sizes of mesh. However, due to the quadratic computation complexity w.r.t. the sequence length of the transformer, it's hard to model all $N=5023$ tokens. Thus, we use the coarse mesh of the FLAME model with $N'=314$ tokens in our vertex-feature transformer and use the decoder of Spiralnet++ \cite{gong2019spiralnet++} to upsample the vertex features after the transformer, which serializes the neighboring vertices based on triangular meshes. 
Our vertex-feature transformer has $6$ transformer encoder layers and the head of MHSA is set to $4$. The feature dimension is set to $C'=128$ and $C=32$. 
Our model is implemented using PyTorch and optimized with the Adam optimizer \cite{kingma2014adam} for a duration of $200$ epochs. The learning rate is set to $1e-4$ and the batch size is set to $16$. 
$\lambda_{L1}$, $\lambda_{vgg}$, and $\lambda_{seg}$ are set to $1.0$, and $\lambda_{id}$ and $\lambda_{a}$ are set to $0.1$.

\paragraph{Metrics}
Following previous works \cite{khakhulin2022realistic}, we evaluate our CVTHead on both self-reenactment and cross-identity reenactment. 
In self-reenactment, the source and driving image come from the same video. In this scenario, the driving image can be viewed as the ground truth. We use the following metrics to measure the reconstruction quality between the driving image and the synthesized results: (1)L1 loss on the masked region; (2) peak signal-to-noise ratio (PSNR); (3) learned perceptual image patch similarity (LPIPS) with pre-trained AlexNet \cite{zhang2018unreasonable}, and (4) multi-scale structured similarity (MS-SSIM). 
In the cross-identity reenactment, the source and driven image come from different subjects. Given the source image of one subject, We random sample a different subject in the validation set as the driving image. This evaluation requires  the model to fully disentangle the identity and expression information. Since ground truth is unavailable, this task can only be evaluated by some proxy metrics. In detail, we use (1) FID \cite{heusel2017gans} to evaluate the image realism; (2) CSIM \cite{zakharov2019few}, which measures the cosine similarity of the identity embeddings from a pre-trained model between the source image and the synthesized image; and (3) image quality assessment (IQA) \cite{su2020blindly}

\begin{figure}[!ht]
    \centering
    \includegraphics[width=0.9 \linewidth]{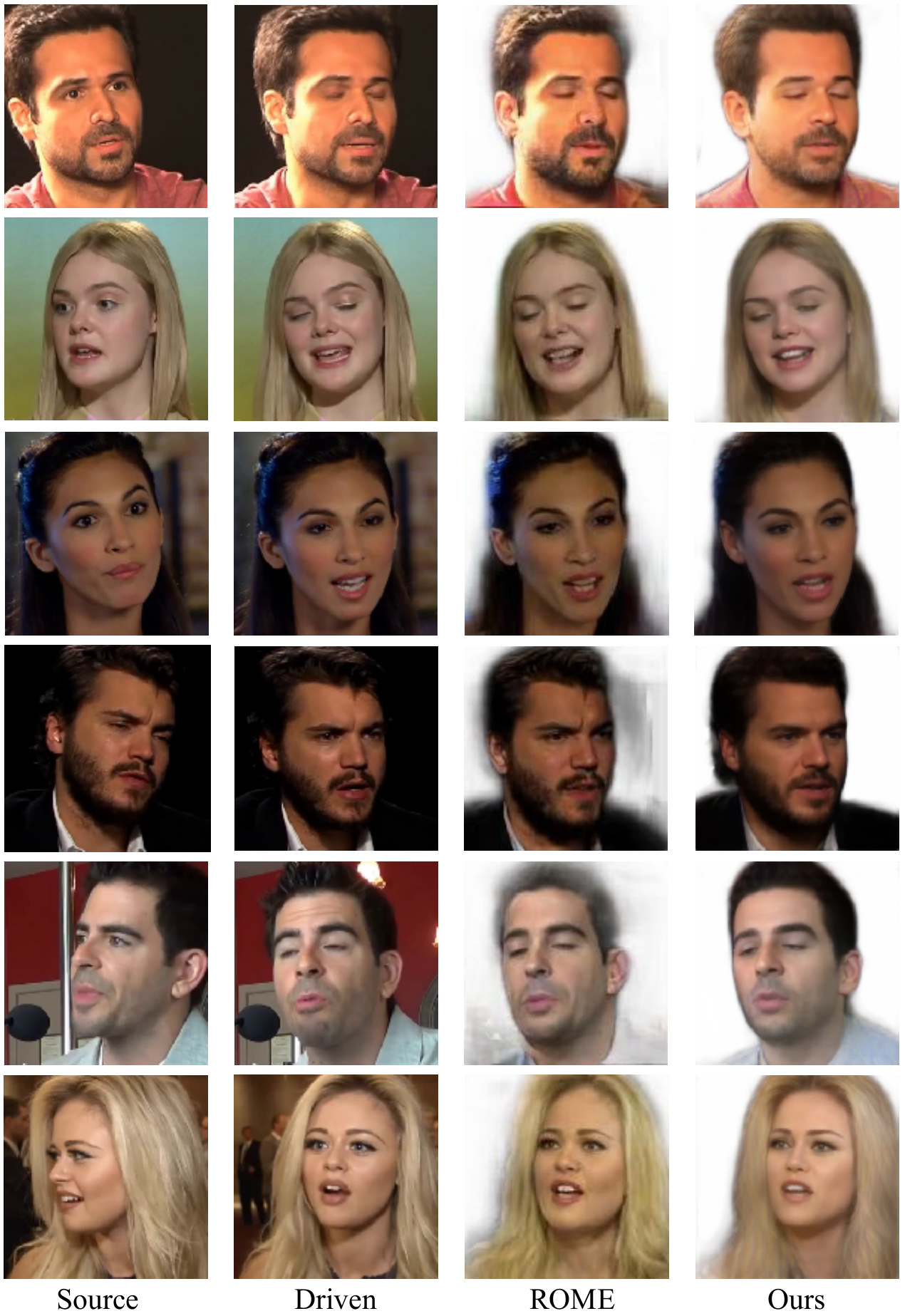}
    \caption{Qualitative comparisons of self-reenactment on VoxCeleb1. The 1st column is the source image. The 2nd column is the driving image, which can be considered as the ground truth. The 3rd column is the results from ROME, and the 4th column is the result from our CVTHead. }
    \label{fig:self_reco}
\end{figure}

\subsection{Results of talking-face synthesis}
We first evaluate the performance of our method on talking-face synthesis. 
To the best of our knowledge, ROME \cite{khakhulin2022realistic} is the only method that share the same setting with our method, i.e., one-shot mesh-based face reenactment based on graphics without warping field. Thus, we mainly compare our method with ROME \cite{khakhulin2022realistic}. Besides, we also compare with warping-based methods including First-Order Motion Model (FOMM) \cite{siarohin2019first} and the Bi-Layer \cite{zakharov2020fast}.

\begin{table}[!ht]

\centering
\resizebox{0.99\linewidth}{!}{
\begin{tabular}{l|c|c|c|c}
\toprule
Dataset & \multicolumn{4}{c}{VoxCeleb1} \\
\midrule
Method & L1 $\downarrow$ & PSNR $\uparrow$ & LPIPS $\downarrow$ & MS-SSIM $\uparrow$ \\
\midrule
FOMM \cite{siarohin2019first} & 0.048 & 22.43 & 0.139 & 0.836 \\
Bi-Layer \cite{zakharov2020fast} & 0.050 & 21.48 & 0.108 & 0.839 \\
ROME \cite{khakhulin2022realistic} & 0.048 & 21.13 & 0.116 & 0.838 \\
Ours & 0.041 & 22.09 & 0.111 & 0.840 \\ 
\midrule

\midrule
Dataset & \multicolumn{4}{c}{VoxCeleb2} \\
\midrule
Method & L1 $\downarrow$ & PSNR $\uparrow$ & LPIPS $\downarrow$ & MS-SSIM $\uparrow$ \\
\midrule
FOMM \cite{siarohin2019first} & 0.059 & 20.93 & 0.165 & 0.793 \\
ROME \cite{khakhulin2022realistic} & 0.050 & 20.75 & 0.117 & 0.834 \\
Ours & 0.042 & 21.37 & 0.119 & 0.841 \\
\bottomrule
\end{tabular}
}
\caption{Results of self-reenactment on the VoxCeleb1 and VoxCeleb2 ($\uparrow$ means larger is better, $\downarrow$ means smaller is better.)}
\label{tab:self_recon}
\end{table}

\paragraph{Self-reenactment}

The quantitative comparison results are summarized in Table \ref{tab:self_recon}.
It is noteworthy that our CVTHead achieves comparable performance with previous methods over all metrics. 
%
Figure \ref{fig:self_reco} illustrates the qualitative comparisons. We also add on the predicted soft mask as in \cite{khakhulin2022realistic} to compare its quality. 
The first three rows showcase scenarios with minimal head rotations and predominantly frontal source images. In such cases, both ROME and  CVTHead exhibit similar performance. However, when the source images depict side views while the driving images present frontal views, ROME tends to generate images with blurry foreground masks in the occluded regions of the source image. Furthermore, ROME often renders these concealed areas in darker colors. These observations indicate that ROME struggles to effectively learn the features of occluded regions and fails to capture the correspondence between mesh vertices. Conversely, our CVTHead addresses these limitations by leveraging transformers to capture long-range dependencies among vertices. These observations suggest that ROME does not effectively learn the features of occluded regions and does not capture the correspondence between mesh vertices. In contrast, our CVTHead addresses these two issues by leveraging transformers to capture long-range dependencies among vertices.

\begin{table}[!ht]
\centering

\resizebox{0.85\linewidth}{!}{
\begin{tabular}{l|c|c|c|c}

\toprule
Dataset & \multicolumn{4}{c}{VoxCeleb1} \\
\midrule
Method & FID $\downarrow$ & CSIM $\uparrow$ & IQA $\uparrow$  & FPS $\uparrow$ \\
\midrule
FOMM \cite{siarohin2019first} & 39.69 & 0.592 & 37.00 &  64.3\\ 
Bi-Layer \cite{zakharov2020fast} & 43.8 & 0.697 & 41.4 & 20.1 \\
ROME \cite{khakhulin2022realistic} & 29.23 & 0.717 & 39.11 & 12.9 \\
Ours & 25.78 & 0.675 & 42.26 & 24.3 \\ 
\midrule

\midrule
Dataset & \multicolumn{4}{c}{VoxCeleb2} \\
\midrule
Method & FID $\downarrow$ & CSIM $\uparrow$ & IQA $\uparrow$  & FPS $\uparrow$ \\
\midrule
FOMM \cite{siarohin2019first} & 61.28 & 0.624 & 36.20 & 64.3 \\
ROME \cite{khakhulin2022realistic} & 53.52 & 0.729 & 37.34 & 12.9 \\ 
Ours & 48.48 & 0.712 & 40.27 & 24.3 \\
\bottomrule
\end{tabular}
}
\caption{Results of cross-identity reenactment. }
\label{tab:cross_recon}
\end{table}

\begin{figure}[!ht]
    \centering
    \includegraphics[width=1.0\linewidth]{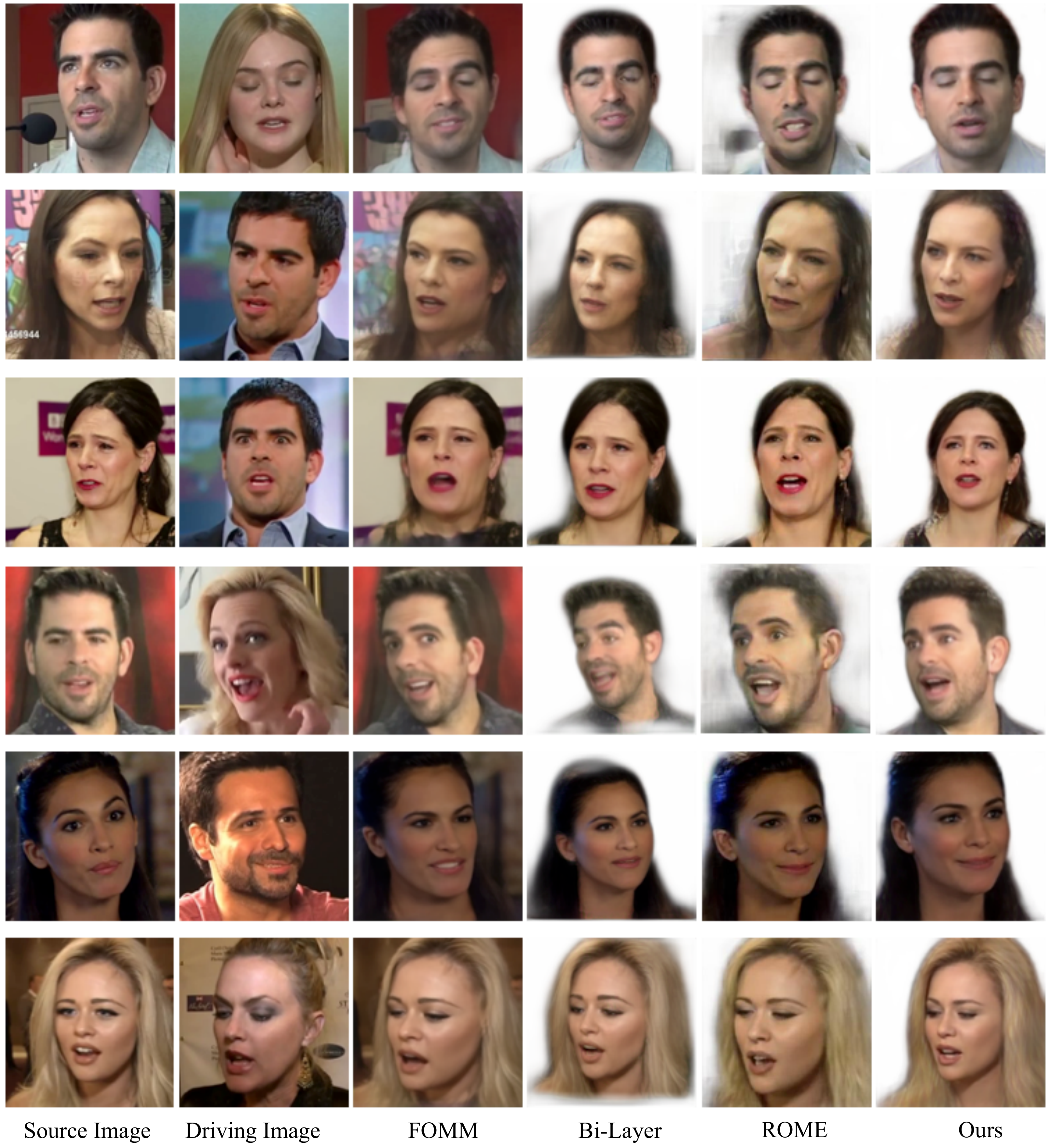}
    \caption{{Qualitative comparisons of cross-identity reenactment on VoxCeleb1.} }
    \label{fig:cross_recon}
\end{figure}

\paragraph{Cross-identity Reenactment}
We proceed to evaluate our method in comparison to other methods for cross-identity reenactment. The quantitative comparison results are presented in Table \ref{tab:cross_recon}. Strikingly, we achieve similar performance on the assessed metrics as ROME, indicating the effectiveness of our method in cross-identity reenactment tasks. 
Furthermore, we provide qualitative results in Figure \ref{fig:cross_recon}, showcasing the ability of our method to generate images with desired expressions, head poses, and other attributes. Notably, warping-based methods usually cannot maintain the identity information such as face shape from the source image. For mesh-guided methods, ROME tends to generate lower-quality images when local regions are occluded in the source image. In contrast, our method demonstrates the capability to maintain the quality of all local regions even in such challenging scenarios.

\begin{figure*}[!ht]
    \centering
    \includegraphics[width=0.95\linewidth]{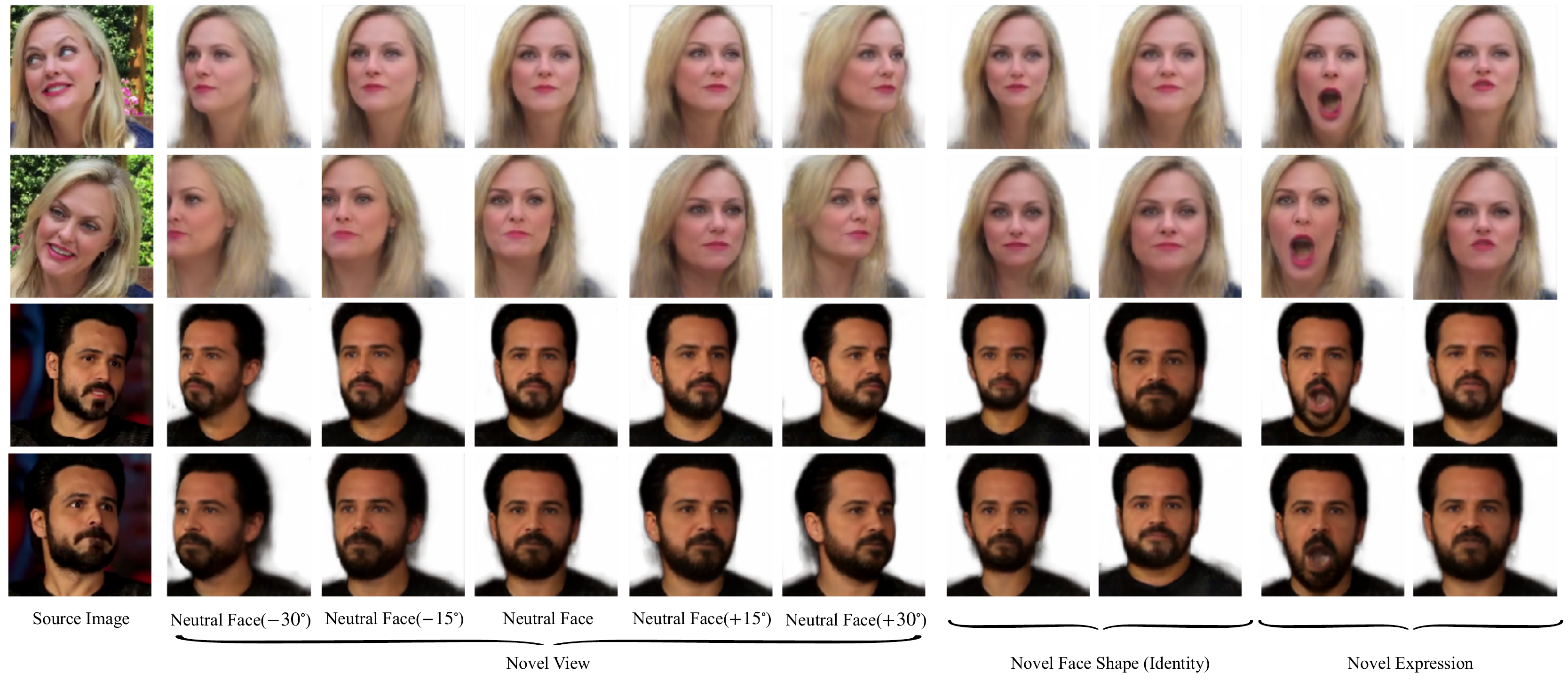}
    \caption{Qualitative results of face animation with novel views, novel face shapes (identity), and novel expressions. }
    \label{fig:face_animation}
\end{figure*}

\paragraph{Inference time comparison}
We also evaluate the inference time of each model, considering the complete duration of 3D mesh reconstruction, the vertex deformation model, and the rendering process. To provide a comprehensive analysis, we report the average FPS (Frames per Second) based on 1000 runs performed on a single RTX 3090Ti. The results are presented in the last column of Table \ref{tab:cross_recon}. Notably, warping-based method is more efficient as they don't need the tedious rendering and mesh reconstruction. 
ROME achieves a modest 12.9 FPS, while our CVTHead model achieves a significantly higher rate of 24.3 FPS. This outcome highlights the superior efficiency of the point-based neural rendering approach compared to traditional graphic-based rendering methods.

\subsection{Results of 3DMM-based Face Animation}

After obtaining the vertex descriptors using the vertex feature transformer, the resulting face can be further manipulated by adjusting the coefficients of the FLAME model \cite{li2017learning}, which control expression $\phi$, pose $\theta$, face shape $\beta$, and camera views $c$. The ability to explicitly control these coefficients enables us to generate faces of the same subject with different expressions, face shapes, and camera views, as illustrated in Figure \ref{fig:face_animation}. This result demonstrates that the learned feature descriptors exhibit a strong alignment with the vertices in the canonical space. Consequently, neural point-based rendering can serve as a viable alternative to traditional graphic-based rendering methods.
Moreover, we intentionally select two distinct source images of the same subject. Interestingly, the generated images, utilizing vertex features from these distinct sources, exhibit a striking resemblance. This intriguing observation further underscores the effectiveness and robustness of our method.

\subsection{Ablation Studies}

\paragraph{Vertex deformation}

\begin{figure}[ht]
    \centering
    \includegraphics[width=1.0\linewidth]{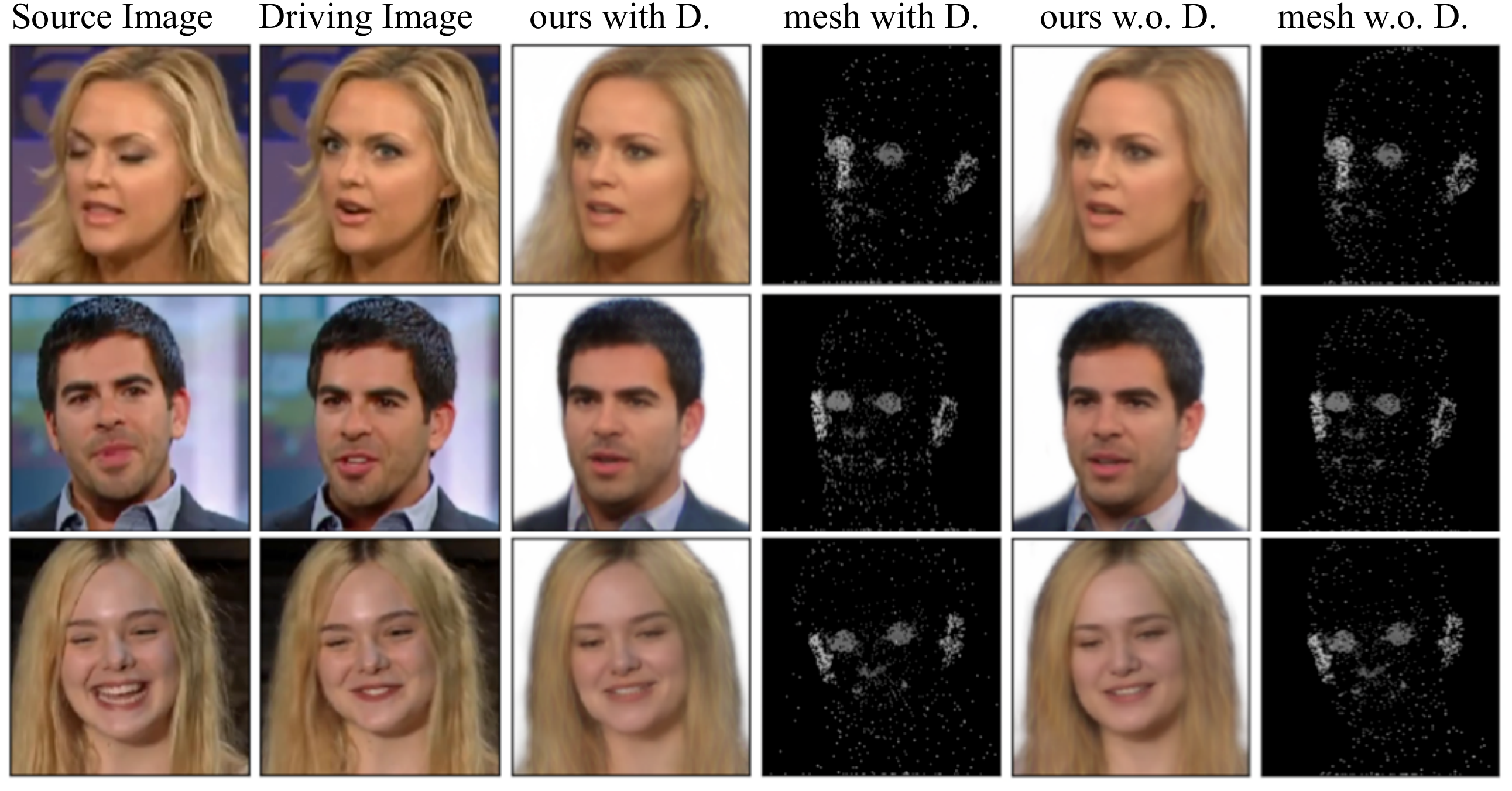}
    \caption{\small{Ablation study of CVTHead with and without the vertex deformation model for hair and shoulder region (D. in short) }}
    \label{fig:abla_deform}
\end{figure}

We utilize the linear deformation model $f_H(\cdot)$ from ROME \cite{khakhulin2022realistic} to deform the vertices of the hair and shoulder region. In this study, we conduct an ablation experiment where we train CVTHead without this vertex deformation module, instead employing the default FLAME mesh with a bald head. 
The results presented in Table \ref{tab:linear_deform} demonstrate that the removal of the vertex deformation (``D." in short) has only a minor impact on the performance. Interestingly, Figure \ref{fig:abla_deform} reveals that the synthesized images from CVTHead, both with and without vertex deformation, appear nearly identical. 
Furthermore, even in cases where the subject has fluffy or long hair that extends beyond the head area, the absence of vertex deformation in CVTHead does not hinder its ability to generate the correct hairstyle. 
These results indicate that the local vertex descriptor can effectively capture the necessary features.

\begin{table}[!ht]
\centering
\resizebox{1.0\linewidth}{!}{
\begin{tabular}{l|c|c|c|c}
\toprule
Method & L1 $\downarrow$ & PSNR $\uparrow$ & LPIPS $\downarrow$ & MS-SSIM $\uparrow$ \\
\midrule
CVTHead (w/o D.) & 0.041 & 22.47 & 0.121 & 0.842 \\
CVTHead & 0.041 & 22.09 & 0.111 & 0.840 \\
\bottomrule
\end{tabular}
}
\caption{\small{Ablation study on the vertex deformation module. We evaluate the performance of self-reenactment on the VoxCeleb1.}}
\label{tab:linear_deform}
\end{table}

\begin{table}[!ht]
\centering
\resizebox{1.0\linewidth}{!}{
\begin{tabular}{l|c|c|c|c}
\toprule
Method & L1 $\downarrow$ & PSNR $\uparrow$ & LPIPS $\downarrow$ & MS-SSIM $\uparrow$ \\
\midrule
Pixel-aligned features & 0.045 & 21.81 & 0.107 & 0.841 \\
CVTHead & 0.041 & 22.09 & 0.111 & 0.840 \\
\bottomrule
\end{tabular}
}
\caption{{Ablation study on the pixel-aligned features}}
\label{tab:pix_align}
\end{table}

\begin{figure}[ht]
    \centering
    \includegraphics[width=0.9\linewidth]{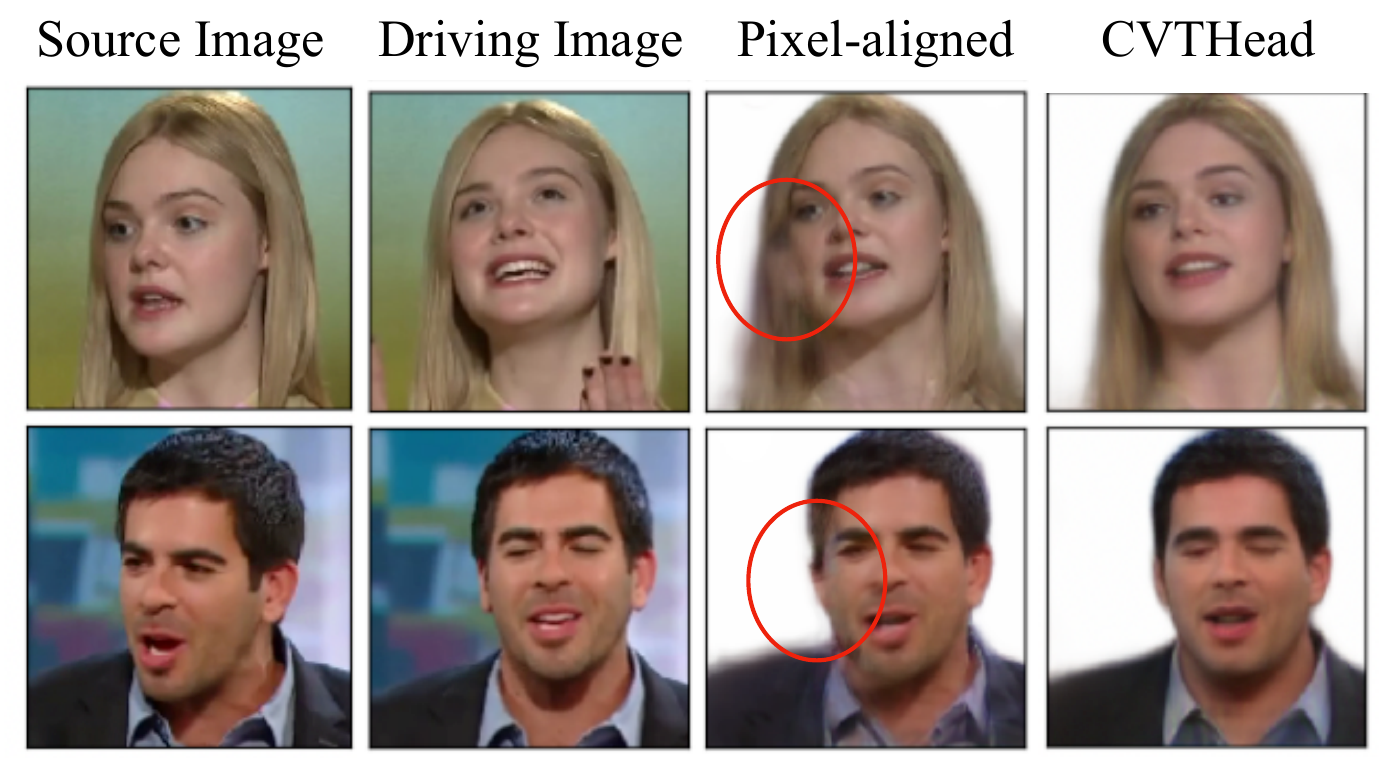}
    \caption{Ablation study of vertex features.}
    \label{fig:abla_s3f}
\end{figure}

\paragraph{Pixel-aligned features}
In our work, we design the vertex feature transformers to learn the vertex feature. In this study, we consider the pixel-aligned features as the baseline, which project the 3D vertex into 2D and choose the corresponding pixel from the image. We follow the architecture design in S3F \cite{corona2023structured} and use a UNet-like feature extractor and sample features of each vertex with its corresponding 2D projection. 
Table \ref{tab:pix_align} indicates that this approach yields a marginally lower PSNR, but a slightly improved LPIPS score. 
As shown in Figure \ref{fig:abla_s3f}, this design can maintain more detailed local features such as hair due to the high-resolution features. Thus, a slightly better LPIPS is achieved. However, when the point is occluded in the source image, the synthesized image tends to generate blur and shadow in these areas if they are visible in the driving pose, which is the reason of the worse PSNR. 
These results suggest that pixel-aligned methods cannot capture the correct features due to the ambiguity of depth. In this case, when a large head rotation happens, this method encounters the same issue as warp-based methods. 

\vspace{-0.5em}
\section{Limitations}
\vspace{-0.5em}
While our method demonstrates effective face animation capabilities from a single image, one potential limitation is that the performance of our approach heavily relies on the accuracy of the 3D mesh reconstruction, specifically utilizing DECA \cite{feng2021learning} in our setup. In certain challenging scenarios, DECA may struggle to fully disentangle the shape and expression factors from the driving images. Consequently, CVTHead may generate images that differ in expressions or head poses from the intended outcome. This highlights the need for further advancements in the accuracy and robustness of 3D mesh reconstruction techniques to address such limitations.

\vspace{-0.5em}
\section{Conclusion}
\vspace{-0.5em}

In this paper, we propose a novel approach for generating explicitly controllable head avatars from a single reference image, utilizing point-based neural rendering. We treat the sparse vertices of the head mesh as a point set and leverage the vertex-feature transformer to learn the local feature descriptor for each vertex. Through our research, we demonstrate that point-based rendering can effectively replace traditional graphic-based rendering methods, offering enhanced efficiency. 
Moreover, we envision that our method can be seamlessly integrated with various generative tools, such as diffusions, to further enhance the quality of generated images and we consider this as  future work.

{\small
\bibliographystyle{ieee_fullname}
\bibliography{egbib}
}

\end{document}